\documentclass[pdflatex,sn-basic]{sn-jnl}
\usepackage{graphicx}%
\usepackage{amsmath,amssymb,amsfonts}%
\usepackage{amsthm}%
\usepackage{mathrsfs}%
\usepackage[title]{appendix}%
\usepackage{xcolor}%
\usepackage{textcomp}%
\usepackage{manyfoot}%
\usepackage{booktabs}%
\usepackage{algorithm}%
\usepackage{algorithmicx}%
\usepackage{algpseudocode}%
\usepackage{listings}%
\usepackage{float}%
\usepackage{hyperref}

\begin{document}

\title[ATV-Net]{ATV-Net: An Adaptive Triple-View Network with Receptive-Field Selection for Efficient Semantic Segmentation}

\author[1]{\fnm{Sheng-Wei} \sur{Chan}$^{\href{https://orcid.org/0009-0002-3983-5163}{[ORCID]}}$}
\author[1]{\fnm{Hsin-Jui} \sur{Pan}}
\author[1]{\fnm{Chun-Po} \sur{Shen}}
\author[1]{\fnm{Chia-Min} \sur{Lin}}
\author[1]{\fnm{Yung-Che} \sur{Wang}}
\author*[1]{\fnm{Jen-Shiun} \sur{Chiang}}\email{chiang@mail.tku.edu.tw}

\affil*[1]{\orgdiv{Department of Electrical and Computer Engineering}, \orgname{Tamkang University}, \orgaddress{\street{No.151, Yingzhuan Rd.}, \city{Tamsui Dist., New Taipei City}, \postcode{251301}, \country{Taiwan}}}

\abstract{%
Recent advances in semantic segmentation rely heavily on attention-based and transformer-style architectures that, while accurate, introduce considerable architectural complexity and computational cost. This paper asks whether a compact CNN-based segmentation head can remain competitive by adaptively selecting useful receptive-field evidence. We propose ATV-Net, an Adaptive Triple-View Network that attaches a lightweight head to a conventional backbone. The head organizes three complementary views---point-wise, neighborhood-level, and enlarged context---and fuses them through an Adaptive Decision Gate that generates image-dependent weights from global feature statistics. This allows the model to emphasize different receptive-field responses according to scene content, without dense attention or multi-scale aggregation. Experiments on Cityscapes and Pascal VOC 2012 show that ATV-Net achieves 80.31\% mIoU on Cityscapes with ResNet-101 and 80.90\% with ConvNeXt-Tiny, and 86.7\% and 88.5\% mIoU on Pascal VOC 2012, respectively, while requiring fewer GFLOPs than representative context-aggregation and attention-based heads. The results indicate that adaptive receptive-field selection remains a practical and effective design choice for CNN-based semantic segmentation.
}


\maketitle
\section{Introduction}
Semantic segmentation is a fundamental task in computer vision that assigns a semantic label to every pixel in an image. It underpins many dense scene-understanding applications, including autonomous driving, intelligent transportation, robotic perception, and urban analysis. In high-resolution street-view scenes, segmentation models must simultaneously recognize large semantic regions such as roads, buildings, vegetation, and sky, and small or thin objects such as poles, traffic signs, pedestrians, riders, and traffic lights. This dual requirement makes segmentation sensitive to both local boundary details and broader contextual information.

Early CNN-based segmentation methods laid the foundation of modern dense prediction. Fully convolutional networks converted image classification backbones into pixel-level predictors~\citep{long2015fcn}, and encoder-decoder models such as U-Net and SegNet emphasized the importance of recovering spatial details~\citep{ronneberger2015unet,badrinarayanan2017segnet}. Subsequent advances in dilated convolution, multi-path refinement, and high-resolution representation~\citep{yu2016dilated,lin2017refinenet,chen2017deeplabv3,chen2018deeplabv3plus,sun2019hrnet} further improved spatial preservation and contextual modeling. Classical context-aggregation modules including PSPNet~\citep{zhao2017pspnet}, DeepLabv3~\citep{chen2017deeplabv3}, DenseASPP~\citep{yang2018denseaspp}, and OCRNet~\citep{yuan2020ocrnet} confirmed that multi-scale and object-level context are essential for robust dense prediction. More recently, attention-based and transformer-based segmentation models have advanced accuracy further by introducing global or long-range feature interaction~\citep{fu2019danet,huang2019ccnet,zheng2021setr,strudel2021segmenter,xie2021segformer,cheng2022mask2former}, while modern ConvNet designs such as ConvNeXt and SegNeXt~\citep{liu2022convnext,guo2022segnext} demonstrate that carefully designed convolutional architectures remain competitive in the transformer era.

These developments raise a practical question. As accuracy gains are increasingly obtained through heavier attention modules, larger pretraining corpora, and transformer-style representation learning, is there still value in a compact CNN-based segmentation head? For settings where architectural simplicity, implementation accessibility, and computational efficiency are first-class concerns, the answer matters. This paper argues that, with the right design, the answer is yes.

Our central observation is that many useful segmentation cues can be captured with elementary receptive-field patterns. A point-wise projection performs reliable semantic transformation; a small local convolution captures boundary and neighborhood structure; an enlarged receptive field provides broader contextual cues. However, combining these responses with a fixed fusion rule is suboptimal: the relative usefulness of point-wise, local, and contextual evidence varies substantially across scenes. Highway frames tend to rely more on enlarged context, while crowded pedestrian scenes need stronger local boundary cues. An adaptive head that selects among such receptive-field views, conditioned on image-level statistics, can therefore recover much of the accuracy gap to heavier context modules at a fraction of their cost.

Motivated by this observation, we propose ATV-Net, an Adaptive Triple-View Network. ATV-Net attaches a lightweight Triple-View Head to a conventional backbone, comprising three complementary branches: a \emph{micro view} that performs point-wise semantic projection, a \emph{local view} that models neighborhood and boundary-sensitive structure, and a \emph{scout view} that captures enlarged contextual cues via dilated convolution. An Adaptive Decision Gate produces image-dependent weights for the three views from a global feature descriptor, replacing static branch concatenation with image-level adaptive selection. A compact Global Coordination Layer then refines the fused representation before the final classifier. The design deliberately avoids dense self-attention and heavy multi-scale aggregation, positioning ATV-Net as a compact adaptive alternative to more complex segmentation heads.

The contributions of this paper are as follows:
\begin{itemize}
    \item We revisit compact CNN-based semantic segmentation and study whether competitive accuracy can be recovered through adaptive receptive-field selection rather than heavier attention or transformer-based modeling.
    \item We propose ATV-Net, an Adaptive Triple-View Network whose Triple-View Head organizes point-wise, local, and enlarged receptive-field responses into a unified and semantically interpretable head.
    \item We introduce an Adaptive Decision Gate that learns image-dependent fusion weights for the three views from global feature statistics, providing a markedly lighter alternative to multi-scale attention and dense context aggregation.
    \item Experiments on Cityscapes and Pascal VOC 2012 show that ATV-Net achieves competitive accuracy across two distinct backbones (ResNet-101 and ConvNeXt-Tiny) and two benchmarks, while requiring fewer GFLOPs than several representative context-aggregation and attention-based heads.
\end{itemize}

\section{Related Work}

\subsection{CNN-Based Semantic Segmentation}

CNN-based semantic segmentation has progressed from fully convolutional prediction to increasingly stronger dense-prediction frameworks. FCN~\citep{long2015fcn} showed that classification networks can be converted into pixel-level predictors. U-Net~\citep{ronneberger2015unet}, SegNet~\citep{badrinarayanan2017segnet}, and RefineNet~\citep{lin2017refinenet} highlighted the value of encoder-decoder recovery, skip connections, and multi-level refinement. Later approaches improved this formulation through dilated convolution, large-kernel context modeling, and stronger backbones~\citep{yu2016dilated,peng2017gcn,chen2017deeplabv3,chen2018deeplabv3plus}. A central challenge is balancing semantic abstraction and spatial precision: deeper layers provide stronger semantics but lower resolution, while shallower features preserve detail but contain weaker semantics. High-resolution designs such as HRNet~\citep{sun2019hrnet} address this by maintaining detailed feature maps throughout the network, while DeepLab-style models use atrous convolution and decoder refinement to preserve dense prediction quality.

ATV-Net follows this CNN-based direction, but its focus is the segmentation head rather than the backbone. It studies how a compact head can improve an existing backbone by organizing complementary receptive-field responses and selecting them adaptively.

\subsection{Multi-Scale Context Aggregation}

Multi-scale context aggregation is important because scene structures appear at different scales. PSPNet~\citep{zhao2017pspnet} introduced pyramid pooling to gather context from multiple spatial regions, while DeepLabv3~\citep{chen2017deeplabv3} used atrous spatial pyramid pooling (ASPP) to capture features with multiple effective receptive fields. GCN~\citep{peng2017gcn} enlarged the effective receptive field with large kernels, and DenseASPP~\citep{yang2018denseaspp} densely connected atrous convolutions to cover a broader scale range. OCRNet~\citep{yuan2020ocrnet} further enhanced semantic consistency through object-contextual representation. Together these methods confirm that contextual cues are essential for segmentation.

Most context modules, however, aggregate multi-scale responses with \emph{static} designs: once the architecture is fixed, the branches are fused by concatenation followed by a $1\times1$ projection, regardless of input content. ATV-Net departs from this paradigm by treating receptive-field fusion as an \emph{adaptive selection problem}. The proposed gate learns image-dependent weights for micro, local, and scout views, allowing the head to emphasize different visual evidence for different scene contents.

\subsection{Attention and Transformer-Based Segmentation}

Attention mechanisms have been widely used to model long-range dependencies in segmentation. DANet~\citep{fu2019danet} introduced dual attention for spatial and channel dependencies, while CCNet~\citep{huang2019ccnet} used criss-cross attention to reduce the cost of dense contextual aggregation. Transformer-based approaches extend this direction further: SETR~\citep{zheng2021setr} reformulated segmentation as a sequence-to-sequence prediction problem; Segmenter~\citep{strudel2021segmenter} employed transformer encoders with mask-based decoding; SegFormer~\citep{xie2021segformer} combined a hierarchical transformer encoder with a lightweight decoder; and Mask2Former~\citep{cheng2022mask2former} used masked attention for unified segmentation tasks.

These methods reflect a different design philosophy from ATV-Net. Rather than competing with large transformer-based frameworks on absolute accuracy, ATV-Net asks whether a lightweight receptive-field selection head can offer a favorable trade-off between accuracy and architectural complexity, particularly for settings constrained by compute or implementation budget.

\subsection{Efficient Segmentation and Dynamic Feature Fusion}

Efficiency-oriented segmentation studies how to obtain useful dense prediction with reduced complexity. ENet~\citep{paszke2016enet}, ERFNet~\citep{romera2018erfnet}, ICNet~\citep{zhao2018icnet}, BiSeNet~\citep{yu2018bisenet}, Fast-SCNN~\citep{poudel2019fastscnn}, DDRNet~\citep{hong2021ddrnet}, STDC~\citep{fan2021stdc}, and PIDNet~\citep{xu2023pidnet} all explore lightweight or real-time designs. These works underscore the importance of balancing accuracy with computational cost.

Dynamic feature fusion is another closely related line. Channel recalibration and selective kernel mechanisms have been explored in SENet~\citep{hu2018senet}, CBAM~\citep{woo2018cbam}, ECA-Net~\citep{wang2020ecanet}, and SKNet~\citep{li2019sknet}, while conditional and dynamic convolution designs adapt model behavior to input content~\citep{yang2019condconv,chen2020dynamicconv}. ATV-Net belongs to this broad direction but adopts a deliberately compact, view-level fusion strategy: rather than applying dense attention or dynamic kernels throughout the network, it summarizes the head's feature map once and emits only three weights --- one for each of the micro, local, and scout views.

\subsection{Positioning of ATV-Net}

Although ATV-Net shares conceptual elements with prior work, its design occupies a distinct point in the design space, summarized below:

\begin{itemize}
\item \textbf{Versus ASPP-style aggregation.} ASPP-based modules~\citep{chen2017deeplabv3,yang2018denseaspp} gather multi-scale context via parallel dilated convolutions followed by \emph{static concatenation and a $1\times1$ projection}. ATV-Net instead treats each branch as a semantically distinct view --- point-wise semantics, local geometry, and enlarged context --- and applies \emph{input-dependent weighting} rather than a fixed projection. The number of branches is also kept deliberately small (three), reducing both parameter and FLOP cost compared with ASPP variants.
\item \textbf{Versus SKNet.} SKNet~\citep{li2019sknet} performs \emph{kernel-level} selection within residual blocks during backbone feature extraction. ATV-Net applies selection at the \emph{segmentation head}, where the choice of receptive field directly affects pixel-wise predictions. The granularity (whole view rather than single kernel), the architectural location (head rather than backbone block), and the downstream task (dense prediction rather than classification) are different.
\item \textbf{Versus SE-style attention.} Squeeze-and-Excitation~\citep{hu2018senet} recalibrates channels within a single feature path. The Adaptive Decision Gate instead produces \emph{cross-branch} fusion weights over different receptive-field views, which is functionally distinct from per-channel modulation within one path. SE-style channel modulation does appear in our pipeline, but only inside the Global Coordination Layer as a post-fusion refinement; it is not the primary fusion mechanism.
\end{itemize}

\section{Method}

\subsection{Overview}

\begin{figure}[t]
\centering
\includegraphics[width=0.82\textwidth]{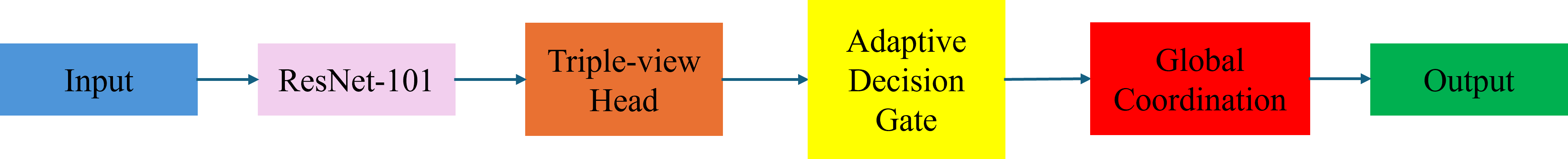}
\caption{Overall framework of ATV-Net. The head uses three simple receptive-field views and an Adaptive Decision Gate to perform input-dependent view fusion, followed by a compact Global Coordination Layer.}
\label{fig:struc}
\end{figure}

The overall architecture of ATV-Net is illustrated in Fig.~\ref{fig:struc}. ATV-Net is a compact adaptive semantic segmentation framework that can be attached to different backbones. Given an input image, the backbone first extracts dense feature representations. The resulting feature map is then processed by a Triple-View Head, which generates three complementary receptive-field views for semantic reasoning. The micro view focuses on point-wise semantic transformation, the local view captures neighborhood and boundary-sensitive structures, and the scout view provides enlarged contextual cues. These views are dynamically fused by the proposed Adaptive Decision Gate. A compact Global Coordination Layer then improves spatial and channel consistency before the final segmentation prediction.

The overall process can be described as:
\begin{equation}
F = B(I),
\end{equation}
where $I$ is the input image and $B(\cdot)$ denotes the backbone network. The feature map $F$ is processed by three view branches:
\begin{equation}
F_m = V_m(F), \quad F_l = V_l(F), \quad F_s = V_s(F),
\end{equation}
where $F_m$, $F_l$, and $F_s$ denote the micro, local, and scout views. The fused triple-view feature is computed as:
\begin{equation}
F_{tv} = \alpha_m F_m + \alpha_l F_l + \alpha_s F_s,
\end{equation}
where $\alpha_m$, $\alpha_l$, and $\alpha_s$ are image-dependent weights generated by the Adaptive Decision Gate, satisfying $\alpha_m + \alpha_l + \alpha_s = 1$.

\subsection{Backbone Feature Extraction}

ATV-Net is evaluated with two backbone settings: ResNet-101~\citep{he2016resnet} with output stride 8 and ConvNeXt-Tiny~\citep{liu2022convnext}. For ResNet-101, the stride-2 operations in the last two residual stages (stages 4 and 5) are replaced with dilated convolutions, so that the final feature map preserves dense spatial information at output stride 8 rather than the default output stride 32. Given an input image $I \in \mathbb{R}^{H \times W \times 3}$, the backbone produces:
\begin{equation}
F \in \mathbb{R}^{\frac{H}{8} \times \frac{W}{8} \times C}.
\end{equation}
Maintaining a dense feature map is important because small objects and object boundaries can be lost after excessive downsampling. Evaluating ATV-Net on two backbones also helps verify that the proposed head is not tied to a specific feature extractor.

\subsection{Triple-View Head}

The Triple-View Head is motivated by the observation that different segmentation cues benefit from different receptive fields. Instead of relying on a single segmentation head or a heavy context module, ATV-Net constructs three simple but complementary views. All three branches project the backbone feature map to a common channel dimension $C'$ so that they can be combined directly by weighted summation.

\paragraph{Micro View.}
The micro view uses a $1 \times 1$ convolution:
\begin{equation}
F_m = \mathrm{Conv}_{1\times1}(F).
\end{equation}
This branch performs point-wise feature projection and focuses on semantic transformation at each spatial location. Because it does not expand the spatial receptive field, it preserves local identity and provides a stable, classification-oriented representation.

\paragraph{Local View.}
The local view uses a $3 \times 3$ convolution:
\begin{equation}
F_l = \mathrm{Conv}_{3\times3}(F).
\end{equation}
This branch captures neighborhood structures, edges, contours, and small geometric patterns. It is particularly useful for refining boundaries and distinguishing adjacent categories with similar appearance.

\paragraph{Scout View.}
The scout view uses a dilated $5 \times 5$ convolution with dilation rate $d$:
\begin{equation}
F_s = \mathrm{Conv}_{5\times5}^{d}(F).
\end{equation}
This branch provides a larger effective receptive field and captures broader contextual cues, helping the model reason about distant objects, partial occlusion, and scene-level relationships.

Together, the three views provide point-wise semantics, local geometry, and enlarged context. The design is intentionally compact: the contribution does not lie in using complex branch operations, but in arranging elementary receptive fields as semantically interpretable views and selecting among them adaptively.

\subsection{Adaptive Decision Gate}

A fixed fusion strategy assumes that the three views should contribute identically across all inputs. This assumption is restrictive because the relative usefulness of point-wise, local, and contextual evidence varies considerably across scenes --- for example, highway frames typically benefit more from enlarged context, while crowded urban scenes rely more on local boundary structure. ATV-Net therefore introduces an Adaptive Decision Gate that generates \emph{image-level} fusion weights from global feature statistics. We deliberately use a single global descriptor rather than a spatially varying gate to keep the head compact; finer-grained region-level gating is left as future work and discussed in Section~\ref{sec:discussion}.

The gate first applies global average pooling to the backbone feature map:
\begin{equation}
z = \mathrm{GAP}(F),
\end{equation}
yielding a single image-level descriptor $z \in \mathbb{R}^{C}$. The descriptor is passed through two compact $1 \times 1$ convolutional layers separated by a ReLU activation $\sigma(\cdot)$, and a softmax produces three normalized weights:
\begin{equation}
\alpha = \mathrm{Softmax}\!\left(\mathrm{Conv}_{1\times1}^{(2)}\!\left(\sigma\!\left(\mathrm{Conv}_{1\times1}^{(1)}(z)\right)\right)\right),
\end{equation}
where the first $1\times1$ convolution reduces the channel dimension from $C$ to $C/r$ (with reduction ratio $r$) for efficiency, and the second $1\times1$ convolution projects the result to a $3$-dimensional vector on which the softmax is applied:
\begin{equation}
\alpha = [\alpha_m, \alpha_l, \alpha_s] \in \mathbb{R}^{3}, \qquad \alpha_m + \alpha_l + \alpha_s = 1.
\end{equation}
The fused triple-view feature is:
\begin{equation}
F_{tv} = \alpha_m F_m + \alpha_l F_l + \alpha_s F_s.
\end{equation}

This design lets ATV-Net select receptive-field responses on a per-image basis. When local structure dominates, the gate emphasizes the local view; when broader context is needed, it shifts weight to the scout view; when point-wise semantics suffice, it relies more on the micro view. Compared with concatenation-based fusion used in ASPP-style modules~\citep{chen2017deeplabv3,yang2018denseaspp}, the gate replaces a fixed projection with an input-conditioned linear combination over three semantically distinct views, while keeping the parameter overhead negligible.

\subsection{Global Coordination Layer}

After adaptive fusion, ATV-Net applies a compact refinement stage. Two $3 \times 3$ convolutional layers are first used to reduce inconsistencies among the three view responses and to improve spatial coherence:
\begin{equation}
F_r = \mathrm{Conv}_{3\times3}\!\left(\mathrm{Conv}_{3\times3}(F_{tv})\right).
\end{equation}

A lightweight channel recalibration step then coordinates the refined feature map at the channel level. Global average pooling summarizes $F_r$:
\begin{equation}
g = \mathrm{GAP}(F_r),
\end{equation}
and two compact $1 \times 1$ convolutions produce a channel-wise modulation vector:
\begin{equation}
\beta = \mathrm{Sigmoid}\!\left(\mathrm{Conv}_{1\times1}^{(2)}\!\left(\delta\!\left(\mathrm{Conv}_{1\times1}^{(1)}(g)\right)\right)\right),
\end{equation}
where $\delta(\cdot)$ is a ReLU activation. The coordinated feature is:
\begin{equation}
F_c = F_r \odot \beta + F_r,
\end{equation}
where $\odot$ denotes channel-wise multiplication with broadcasting over spatial dimensions, and the residual connection preserves the unrecalibrated signal. We note that this step is closely related to SE-style channel attention~\citep{hu2018senet}; we use it here as a \emph{post-fusion refinement} rather than as the primary fusion mechanism --- the latter role is played by the Adaptive Decision Gate, which operates on cross-branch view weights rather than channel weights. Finally, $F_c$ is passed to the segmentation classifier and upsampled to the input resolution.

\subsection{Training Objective}

ATV-Net is trained with pixel-wise cross-entropy loss combined with Online Hard Example Mining (OHEM)~\citep{shrivastava2016ohem}, which focuses optimization on pixels with high prediction loss. This is important because easy background pixels dominate segmentation training and can mask gradients from boundaries and small objects. Letting $\Omega_h$ denote the set of selected hard pixels, the OHEM loss is:
\begin{equation}
\mathcal{L}_{\mathrm{OHEM}} = -\frac{1}{|\Omega_h|}\sum_{i \in \Omega_h} y_i \log p_i,
\end{equation}
where $y_i$ is the ground-truth label and $p_i$ is the predicted probability for pixel $i$.

\section{Experiments}

\subsection{Datasets}

We evaluate ATV-Net on Cityscapes~\citep{cordts2016cityscapes} and Pascal VOC 2012~\citep{everingham2015pascal}. Cityscapes is a high-resolution urban scene-parsing benchmark consisting of street-view images with fine pixel-level annotations across 19 evaluation classes. Pascal VOC 2012 is a widely used segmentation benchmark with diverse object categories and scene layouts. For Pascal VOC 2012, following common practice, we use the augmented training set built from the official Pascal VOC 2012 training set and the Semantic Boundaries Dataset (SBD)~\citep{hariharan2011semantic}. Using both datasets lets us assess whether the proposed adaptive head transfers beyond a single urban-scene domain.

\subsection{Evaluation Metric}

We use mean Intersection over Union (mIoU) as the evaluation metric. For each class, IoU is computed as:
\begin{equation}
\mathrm{IoU} = \frac{TP}{TP + FP + FN},
\end{equation}
where $TP$, $FP$, and $FN$ denote true positives, false positives, and false negatives. The final mIoU is the per-class average of IoU.

\subsection{Implementation Details}

ATV-Net is implemented with ResNet-101 and ConvNeXt-Tiny backbones; both are initialized from weights pretrained on ImageNet-1K. For ResNet-101, the output stride is set to 8 to preserve dense spatial features for high-resolution segmentation. For the Triple-View Head, all three view branches project the backbone feature to a common channel dimension of $C' = 256$; the scout view uses a dilation rate of $d = 2$; and the Adaptive Decision Gate uses a reduction ratio of $r = 8$. For Cityscapes training, input images are randomly cropped to $768 \times 768$ and the model is trained with a batch size of 8 for 150 epochs. For Pascal VOC 2012, input images are randomly cropped to $512 \times 512$ with a batch size of 4; all other training settings follow the Cityscapes configuration. Standard data augmentation is applied, including random scaling, random cropping, and horizontal flipping. All experiments are conducted on a single NVIDIA RTX 5080 GPU. We use AdamW~\citep{loshchilov2017decoupled} as the optimizer with an initial learning rate of $1\times10^{-4}$ and a poly learning-rate schedule. The model is trained with the OHEM cross-entropy loss described above. At evaluation, we use single-scale inference without multi-scale or flip testing, so that the reported numbers reflect raw model performance rather than test-time augmentation.

\subsection{Main Results on Cityscapes}
\begin{table}[t]
\centering
\caption{Comparison with representative semantic segmentation methods on the Cityscapes validation set: PSPNet~\citep{zhao2017pspnet}, DeepLabv3~\citep{chen2017deeplabv3}, DeepLabv3+~\citep{chen2018deeplabv3plus}, DANet~\citep{fu2019danet}, CCNet~\citep{huang2019ccnet}, and ConvNeXt-Tiny~\citep{liu2022convnext}. ATV-Net is positioned as a compact adaptive head rather than a large state-of-the-art model.}
\label{tab:main_result}
\footnotesize
\setlength{\tabcolsep}{4pt}
\begin{tabular}{lcccc}
\toprule
Method & Backbone & Main Head / Module & Adaptive Fusion & mIoU (\%) \\
\midrule
PSPNet & ResNet-101 & Pyramid Pooling & No & 78.4 \\
DeepLabv3 & ResNet-101 & ASPP & No & 79.30 \\
DeepLabv3+ & Xception-65 & ASPP + Decoder & No & 79.10 \\
DANet & ResNet-101 & Dual Attention & Yes & 79.93 \\
CCNet & ResNet-101 & Criss-Cross Attention & Yes & 80.50 \\
\midrule
ATV-Net & ResNet-101 & Triple-View Head & Yes & 80.31 \\
ATV-Net & ConvNeXt-Tiny & Triple-View Head & Yes & 80.90 \\
\bottomrule
\end{tabular}
\end{table}

Table~\ref{tab:main_result} reports results on the Cityscapes validation set. ATV-Net achieves 80.31\% mIoU with ResNet-101 and 80.90\% mIoU with ConvNeXt-Tiny. These results show that a compact adaptive receptive-field head can match or come close to several representative CNN-based, context-aggregation, and attention-based heads built on the same backbone, without resorting to dense attention or transformer-style global modeling. They support the central motivation of this work: when receptive-field evidence is selected adaptively, architectural simplicity in the head remains a viable design choice.

\subsection{Backbone and Dataset Generalization}

\begin{table}[t]
\centering
\caption{Generalization of ATV-Net across different backbones and datasets.}
\label{tab:backbone_dataset}
\begin{tabular}{lcc}
\toprule
Backbone & Cityscapes mIoU (\%) & Pascal VOC 2012 mIoU (\%) \\
\midrule
ResNet-101, OS=8 & 80.31 & 86.7 \\
ConvNeXt-Tiny & 80.90 & 88.5 \\
\bottomrule
\end{tabular}
\end{table}

Table~\ref{tab:backbone_dataset} summarizes results across different backbones and datasets. Switching to ConvNeXt-Tiny improves the Cityscapes result to 80.90\% mIoU. On Pascal VOC 2012~\citep{everingham2015pascal}, ATV-Net further obtains 86.7\% mIoU with ResNet-101 and 88.5\% mIoU with ConvNeXt-Tiny. This indicates that ATV-Net is not limited to a single ResNet-based implementation or a single urban-scene dataset; the Pascal VOC result in particular validates the proposed head on more diverse, object-centric scenes.

\subsection{Ablation Study}

\begin{table}[t]
\centering
\caption{Ablation study of receptive-field views and adaptive fusion on the Cityscapes validation set.}
\label{tab:ablation}
\begin{tabular}{lcc}
\toprule
Model Variant & Fusion Strategy & mIoU (\%) \\
\midrule
Micro only & Single view & 76.74 \\
Local only & Single view & 77.46 \\
Scout only & Single view & 77.85 \\
Micro + Local & Fixed fusion & 78.51 \\
Micro + Scout & Fixed fusion & 78.73 \\
Local + Scout & Fixed fusion & 79.20 \\
\midrule
ATV-Net & Adaptive Decision Gate & 80.31 \\
\bottomrule
\end{tabular}
\end{table}

Table~\ref{tab:ablation} verifies the effectiveness of the proposed components. Single-view variants are clearly weaker because each receptive-field view captures only one type of evidence. Two-view variants improve performance, indicating that complementary receptive fields are useful. The full ATV-Net achieves the best result among the evaluated variants, suggesting that combining the three receptive-field views with the Adaptive Decision Gate is beneficial for selecting useful responses according to input scene characteristics. This supports the paper's main argument that simple branches become more effective when adaptively fused rather than statically combined.

\subsection{Computational Complexity}

\begin{table}[t]
\centering
\caption{Computational complexity comparison under different input resolutions. GFLOPs are measured under the listed resolutions.}
\label{tab:complexity}
\begin{tabular}{lcccc}
\toprule
Model & $512 \times 512$ & $512 \times 1024$ & $1024 \times 1024$ & $1024 \times 2048$ \\
\midrule
DeepLabV3+ & 348.0 & 509.0 & 1018.0 & 2035.0 \\
DANet & 289.0 & 597.0 & 1272.0 & 2542.0 \\
CCNet & 276.0 & 551.0 & 1103.0 & 2206.0 \\
PSPNet & 256.0 & 513.0 & 1026.0 & 2053.0 \\
DeepLabV3 & 254.0 & 695.0 & 1391.0 & 2781.0 \\
OCRNet & 325.0 & 652.0 & 1246.0 & 2342.0 \\
\midrule
ATV-Net & 237.0 & 474.0 & 947.0 & 1894.0 \\
\bottomrule
\end{tabular}
\end{table}

ATV-Net contains 49.14M parameters and requires 947 GFLOPs for a $1024 \times 1024$ input. Table~\ref{tab:complexity} reports computational complexity under multiple input resolutions; GFLOPs scale with the input image area as expected. Compared with representative context-aggregation and attention-based segmentation models under the same resolution settings, ATV-Net consistently requires fewer GFLOPs. This indicates that ATV-Net's competitive accuracy does not come from increasing model complexity, but from a compact adaptive head that organizes basic receptive-field responses effectively.

\subsection{Qualitative Analysis}

\begin{figure}[t]
\centering
\includegraphics[width=0.48\textwidth]{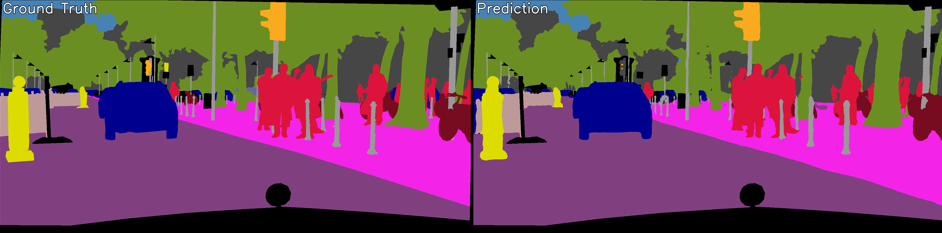}
\hfill
\includegraphics[width=0.48\textwidth]{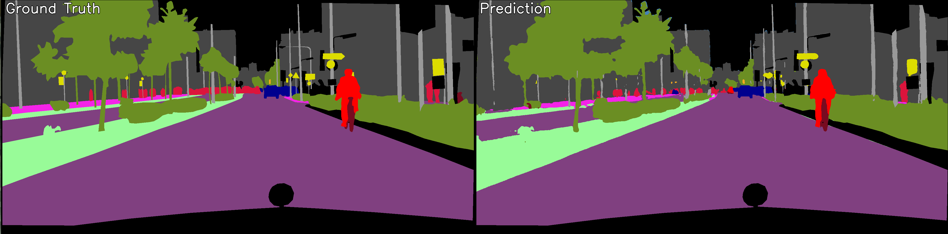}
\caption{Qualitative segmentation results on the Cityscapes validation set. Each pair shows the ground-truth annotation and the prediction generated by ATV-Net.}
\label{fig:qualitative}
\end{figure}

Figure~\ref{fig:qualitative} shows qualitative segmentation results. ATV-Net produces visually coherent predictions while preserving many object boundaries and local structures. These examples are consistent with the motivation of the Triple-View Head: local and enlarged receptive-field responses provide complementary information, while adaptive fusion helps select useful evidence for different scene contents.

\subsection{Discussion}\label{sec:discussion}

The results above let us examine, beyond raw mIoU numbers, what the design of ATV-Net actually achieves and where its boundaries lie.

\paragraph{What the ablation says.}
Table~\ref{tab:ablation} indicates that no single view is sufficient. Each single-view variant lies in a roughly 1\,percentage-point band (76.74\%--77.85\%), confirming that point-wise, local, and contextual cues each carry only part of the segmentation signal. Two-view variants close most of the gap to the full model but still trail the full ATV-Net by 1.1--1.8 percentage points. This pattern is consistent with our hypothesis that the three views are genuinely complementary and that adaptive selection contributes additional value beyond merely combining more views.

\paragraph{Why a global gate rather than a spatial one.}
The Adaptive Decision Gate uses global average pooling to summarize the feature map before producing the view weights. This means a single set of weights $(\alpha_m, \alpha_l, \alpha_s)$ is shared across all spatial locations of a given image. The choice is intentional: it keeps the gate cost negligible relative to the backbone and the view branches, and it captures scene-level priors (urban vs.~highway, dense vs.~sparse) that are typically the most informative cue for choosing among receptive-field views. A per-region or per-pixel gate would be expressively stronger, but would also incur additional parameter and FLOP cost; we view it as a natural next step rather than a property of the current design.

\paragraph{What ATV-Net does not do.}
ATV-Net does not aim to push absolute accuracy beyond modern transformer-based or strongly pretrained frameworks. On Cityscapes, recent transformer-based heads can exceed our numbers by several mIoU points, typically at substantially higher computational and architectural cost. The contribution of this work is therefore better characterized as a \emph{simplicity / efficiency trade-off}: among heads built on top of standard CNN backbones without large-scale pretraining or dense attention, an adaptively gated triple-view head is a competitive option, and a notably compact one.

\paragraph{Limitations.}
We highlight three limitations honestly. First, as noted above, the gate is image-level only; segmentation tasks that demand strongly localized switching between receptive fields (e.g., scenes with both very fine-grained and very large objects) may benefit from a spatially varying gate. Second, all reported numbers use single-scale inference without test-time augmentation; multi-scale or flip testing would likely raise our numbers but is not the focus of this work. Third, our comparison set is restricted to heads of broadly similar architectural class; we do not directly contrast against very large or heavily pretrained transformer segmentation frameworks because they target a different operating point.

\section{Conclusion}

We presented ATV-Net, an Adaptive Triple-View Network for semantic segmentation. ATV-Net organizes elementary receptive-field responses --- point-wise, local, and enlarged --- into three semantically interpretable views, and uses a compact Adaptive Decision Gate to fuse them with image-dependent weights. A lightweight Global Coordination Layer further refines the fused representation before classification. Experiments on Cityscapes and Pascal VOC 2012 show that this compact head delivers competitive accuracy across two distinct backbones and two datasets, while consuming fewer GFLOPs than several representative context-aggregation and attention-based heads. The results suggest that adaptive receptive-field selection remains a practical design lever for CNN-based semantic segmentation, complementary to the broader trend toward heavier attention modules and transformer-style modeling. Promising future directions include region-level gating, real-time variants, and combinations with stronger pretrained backbones.

\backmatter

\section*{Statements and Declarations}

\bmhead{Funding}
This research work was partially supported by the National Science and Technology Council, Taiwan, under grant number 114-2221-E-032-011.

\bmhead{Competing Interests}
The authors declare that they have no competing interests that are relevant to the content of this article.

\bmhead{Ethics approval and consent to participate}
Not applicable.

\bmhead{Consent for publication}
Not applicable.

\bmhead{Data availability}
The datasets used in this study, Cityscapes and Pascal VOC 2012 (with SBD), are publicly available from their respective official sources.

\bmhead{Code availability}
The source code will be made publicly available on GitHub upon publication.

\bmhead{Author contributions}
Following the CRediT taxonomy:
\textbf{Sheng-Wei Chan}: Conceptualization, Methodology, Software, Investigation, Formal analysis, Writing -- original draft.
\textbf{Hsin-Jui Pan}: Investigation, Formal analysis, Writing -- review \& editing.
\textbf{Chun-Po Shen}: Investigation, Validation, Writing -- review \& editing.
\textbf{Chia-Min Lin}: Software, Validation.
\textbf{Yung-Che Wang}: Data curation, Visualization.
\textbf{Jen-Shiun Chiang}: Supervision, Project administration, Funding acquisition, Writing -- review \& editing.
All authors read and approved the final manuscript.


\end{document}